\def\year{2022}\relax
\documentclass[letterpaper]{article} 
\usepackage{aaai22}  
\usepackage{times}  
\usepackage{helvet}  
\usepackage{courier}  
\usepackage[hyphens]{url}  
\usepackage{graphicx} 
\usepackage{natbib}  
\usepackage{caption} 
\DeclareCaptionStyle{ruled}{labelfont=normalfont,labelsep=colon,strut=off} 
\usepackage{algorithm}
\usepackage{algorithmic}
\usepackage{todonotes}
\usepackage{booktabs}
\usepackage{pgfplots}
\usepackage{tikz}
\usepackage{standalone}
\usepackage{numprint}
\presetkeys%
    {todonotes}%
    {inline,backgroundcolor=yellow}{}

\renewcommand\todo[1]{}

\usepackage{newfloat}
\usepackage{listings}
\floatstyle{ruled}
\newfloat{listing}{tb}{lst}{}
\floatname{listing}{Listing}

\nocopyright

\usepackage{amsthm}
\usepackage{xspace}
\usepackage{stmaryrd}
\usepackage{subcaption}

\newcommand\boldred[1]{\textcolor{red}{\textbf{#1}}}
\newcommand\graytext[1]{\textcolor{gray}{#1}}

\title{Scaling up ML-based Black-box Planning with Partial STRIPS Models}
\author {
    Matias Greco\textsuperscript{\rm 1},
    {\'A}lvaro Torralba\textsuperscript{\rm 2},
    Jorge A. Baier\textsuperscript{\rm 1},
    Hector Palacios\textsuperscript{\rm 3}
}
\affiliations {
    \textsuperscript{\rm 1} Pontificia Universidad Cat\'olica de Chile, Chile\\
    \textsuperscript{\rm 2} Aalborg University, Denmark\\
    \textsuperscript{\rm 3} ServiceNow Research, Canada\\
    mogreco@uc.cl,
    alto@cs.aau.dk,
    jabaier@ing.puc.cl,
    hector.palacios@servicenow.com
}

\usepackage{amsmath,amsfonts,bm}

\def\eqref#1{equation~\ref{#1}}

\def\1{\bm{1}}

\DeclareMathAlphabet{\mathsfit}{\encodingdefault}{\sfdefault}{m}{sl}
\SetMathAlphabet{\mathsfit}{bold}{\encodingdefault}{\sfdefault}{bx}{n}

\newcommand{\task}{\ensuremath{\Pi}\xspace}
\newcommand{\bbtask}{\ensuremath{\task_{B}}\xspace}
\newcommand{\ptask}{\ensuremath{\task_{D}}\xspace}
\newcommand{\Facts}{\ensuremath{{F}}\xspace}
\newcommand{\Ops}{\ensuremath{{A}}\xspace}
\newcommand{\Sinit}[0]{\ensuremath{s_I}\xspace}
\newcommand{\Sgoal}[0]{\ensuremath{S_G}\xspace}
\newcommand{\Goal}[0]{\ensuremath{G}\xspace}
\newcommand{\pmOset}[2]{\mathrm{#1}(#2)}
\newcommand{\Opre}[1]{\ensuremath{\pmOset{pre}{#1}}\xspace}
\newcommand{\Oadd}[1]{\ensuremath{\pmOset{add}{#1}}\xspace}
\newcommand{\Odel}[1]{\ensuremath{\pmOset{del}{#1}}\xspace}

\theoremstyle{plain}
\newtheorem{theorem}{Theorem}

\theoremstyle{definition}
\newtheorem{definition}[theorem]{Definition}

\newcommand{\hff}{\ensuremath{h^{\text{FF}}}\xspace}

\newcommand{\hffp}[1]{\ensuremath{\hff_{{#1}}}\xspace}

\newcommand{\logisticsone}{\ensuremath{\text{air}}\xspace}
\newcommand{\logisticsoneshort}{A}

\newcommand{\gridtwo}{\ensuremath{\text{robot}}\xspace}
\newcommand{\gridtwoshort}{R}

\newcommand{\gridthree}{\ensuremath{\text{keys}}\xspace}
\newcommand{\gridthreeshort}{K}

\newcommand{\woodone}{\ensuremath{\text{wood}}\xspace}
\newcommand{\woodoneshort}{W}

\newcommand{\woodtwo}{\ensuremath{\text{logistics}}\xspace}
\newcommand{\woodtwoshort}{L}

\newcommand{\hgn}{\ensuremath{h^{\text{hgn}}}\xspace}
\newcommand{\hgndisc}{\ensuremath{\pi^{\text{hgn}}}\xspace}

\newcommand{\hgnd}[1]{\ensuremath{h^{\text{#1}}}\xspace}
\newcommand{\hgndiscd}[1]{\ensuremath{\pi^{\text{#1}}}\xspace}

\newcommand{\logisticsdumbone}{\ensuremath{\text{hgn-one}}}

\newcommand{\hgnlogdumbone}{\hgnd{\logisticsdumbone}}
\newcommand{\hgndisclogdumbone}{\hgndiscd{\logisticsdumbone}}

\newcommand{\griddumbone}{\ensuremath{\text{hgn-gridsize}}}

\newcommand{\hgngriddumbone}{\hgnd{\griddumbone}}
\newcommand{\hgndiscgriddumbone}{\hgndiscd{\griddumbone}}

\newcommand{\griddumbtwo}{\ensuremath{\text{hgn-onelock}}}

\newcommand{\hgngriddumbtwo}{\hgnd{\griddumbtwo}}
\newcommand{\hgndiscgriddumbtwo}{\hgndiscd{\griddumbtwo}}

\newcommand{\woodworkingdumbone}{\ensuremath{\text{hgn-oneloc}}}

\newcommand{\hgnwooddumbone}{\hgnd{\woodworkingdumbone}}
\newcommand{\hgndiscwooddumbone}{\hgndiscd{\woodworkingdumbone}}

\newcommand{\woodworkingdumbtwo}{\ensuremath{\text{hgn-move}}}

\newcommand{\hgnwooddumbtwo}{\hgnd{\woodworkingdumbtwo}}
\newcommand{\hgndiscwooddumbtwo}{\hgndiscd{\woodworkingdumbtwo}}

\begin{document}

\maketitle

\begin{abstract}
A popular approach for sequential decision-making is to perform simulator-based search
guided with Machine Learning (ML) methods like policy learning.  On the other hand,
model-relaxation heuristics can guide the search effectively if a full declarative model
is available. In this work, we consider how a practitioner can improve ML-based black-box
planning on settings where a complete symbolic model is not available. We show that
specifying an incomplete STRIPS model that describes only part of the problem enables the
use of relaxation heuristics. Our findings on several planning domains suggest that this
is an effective way to improve ML-based black-box planning beyond collecting more data or
tuning ML architectures.
\end{abstract}

\section{Introduction}

\todo{H: Deberiamos definir $\hff_{\logisticsoneshort}$ y otras en el main body. Cree comando $\hffp{\logisticsoneshort}$}

Multiple AI areas deal with sequential decision-making problems. When a black-box simulator is available, an increasingly popular
approach is to use search guided with \emph{learned}
policies~\cite{williams-1992} or value/heuristics functions~\cite{silver-et-al-nature2016}, either by ML
using collected data or by online reinforcement learning (RL)~\cite{sutton-barto-2018}. The main
advantage of these approaches is that the simulator can implement an arbitrarily complex
transition function, as a symbolic model is not required to guide the search. However,
such ML methods sometimes feature high sample complexity or fail to generalize to instances beyond
the training data distribution, running out of time or misleading the search.  For
improving them, practitioners might attempt to collect more data or to tune the ML
algorithms.

An alternative is to use a state-of-the-art automated planner, which leverages a full
symbolic description of the state space, e.g., expressed in the \textit{Planning Domain
Definition Language} (PDDL) \cite{mcdermott-aimag2000}. This results in a symbolic model
(e.g. in STRIPS formalism~\cite{fikes-nilsson-aij1971}) that compactly describes the state
space, and can be used to guide the search via relaxation-based
heuristics~\cite{bonet-geffner-aij2001,hoffmann-nebel-jair2001}. In fact, recent
comparisons have shown that, whenever a declarative description is available, these
model-relaxation heuristics often outperform ML-based heuristics even when learning for a
fixed planning task~\cite{ferber-et-al-ecai2020}.

In this paper, we are interested in settings where plans are obtained by searching using an ML-based method and a full symbolic model of the black-box simulator is not available. Our hypothesis is that
such systems can be enhanced by using an incomplete STRIPS model and combining the
guidance of the learned heuristics/policies  with relaxation-based heuristics (e.g. FF~\cite{hoffmann-nebel-jair2001}) over the
partial, incomplete, model.
We consider situations where a hypothetical practitioner has
already trained a heuristic and/or policy to find sequences of actions that reach a goal.
Then, to improve performance, the practitioner models part of
the domain in STRIPS/PDDL, resulting in an incomplete model that describes only part of
the state-space without assuming that the black-box state is fully symbolic. The question we tackle
is, can the practitioner improve the performance of the overall system by specifying only
part of the model in a symbolic way, while still relying on the ML-guidance for capturing
other (e.g., non-symbolic) aspects?  And if so, how important it is which parts of the
model are specified?

\todo{Previous approach has been demonstrate that combine multiple heuristics gives good results (citar "the more the merrier: Combining Heuristic Estimators for Satisficing Planning" (Gaby Roger and Malte) }

We take a first step towards answering these questions by running several case studies on
planning domains for which a full model is available. This allows us to exhaustively
analyze the performance under a diverse range of partial models that focus on different
aspects of the domain and compare the relative improvement with respect to the best possible scenario: having a full STRIPS model. Moreover, we also
considered cases where the data was the most convenient for training the chosen ML-based baseline. Our
methodology is similar to ML papers that evaluate methods where the data is obtained from
a known generative model.

Our experiments show that partial STRIPS models can indeed improve the performance over
ML-based heuristics/policies that show low coverage when used in isolation. This is the case even when both search guidance mechanisms are combined in a
simple way, not requiring a heavily engineered algorithm nor any additional training,
which could be expensive. While the overall performance improves as models become more
accurate, it sometimes may suffice to specify a small part of the domain to achieve a
noticiable improvement. Therefore, our findings suggest that specifying partial symbolic
models is a good alternative way to improve ML-based black-box planning beyond collecting
more data or tuning ML architectures.


\section{Background / Planning Representation}

A \emph{planning task} is a tuple $\task = \langle S, A, f, \Sinit, \Sgoal \rangle$, where $S$ is a finite set of states, $A$ is a finite set of
actions, $f: S \times A \rightarrow S$ is the transition function, $\Sinit \in S$ is the
initial state and $\Sgoal\subseteq S$ is the set of goal states. A plan $\pi = \langle
a_0, \ldots, a_n \rangle$ is a sequence of actions from $s_0$ to any state $s_g \in
\Sgoal$ such that $f(f(f(s_0, a_0), a_1), \ldots, a_n) = s_g$.

A planning task can be specified in multiple ways. Following \citet{katz-et-al-2018}, a \emph{black-box planning task} is a tuple
$\bbtask = \langle S, A, \Sinit, \mathit{succ}, \mathit{goal} \rangle$, where $\mathit{succ}$
and $\mathit{goal}$ are arbitrary, black-box, functions, which define $f(s, a) =
\mathit{succ} (s, a)$  and $\Sgoal
= \{s_g \mid \mathit{goal}(s) = \mathrm{True}\}$, respectively. Black-box tasks are suitable for
performing lookahead search, by starting at the initial state and repeteadly applying the
$\mathit{succ}$ function. Such a search can be guided using heuristics. Formally,
a heuristic $h : S \rightarrow \mathbb{R}^+_0$ is a function mapping states to numeric
values, such as $h(s)$ estimates the distance from $s$ to the goal.  The heuristic
function can be implemented manually, or, e.g., learned automatically from previous tasks~\cite{ferber-et-al-ecai2020,shen-et-al-icaps2020}.

On the other hand, declarative approaches specify planning tasks by representing states in
terms of a set of variables or fluents~\cite{fikes-nilsson-aij1971,backstrom-nebel-compint1995}. A
STRIPS task is a tuple $\ptask = \langle \Facts, \Ops, \Sinit, \Goal \rangle$, where $\Facts$
is a set of facts, and $\Ops$ is a set of actions. A {state} $s \subseteq \Facts$ is a set of
facts, $\Sinit \subseteq \Facts$ is the {initial state} and $\Goal \subseteq \Facts$ is
the {goal} specification.
The semantics of \ptask are defined as follows. $S = 2^F$ is the set of all states, which
correspond to sets of facts that are true in such a state. $S_G
= \{s_g \mid \Goal \subseteq s_g \}$. An action $a \in A$ is a tuple
$\langle \Opre{a}, \Oadd{a}, \Odel{a} \rangle$, where $\Opre{a} \subseteq \Facts$ is a set
of preconditions, and $\Oadd{a} \subseteq \Facts$ and $\Odel{a} \subseteq \Facts$ are sets
of add and delete effects, respectively.  An action $a$ is {applicable} in a state $s$ if
$\Opre{a} \subseteq s$.  The {resulting state} of applying an applicable action $a$ in a
state $s$ is the state $f(s,a) = (s \setminus \Odel{a}) \cup \Oadd{a}$.

STRIPS models are usually expressed in PDDL, where the set of facts $\Facts$ is the set of all possible grounding of a fixed set of predicates over a fixed set of objects ~\cite{mcdermott-et-al-tr1998}.
Having a STRIPS model enables new ways for automatically obtaining heuristic functions via
domain-independent relaxations of the problem, such as the \emph{delete-relaxation} FF
heuristic~\cite{hoffmann-nebel-jair2001}.

\section{Guiding Black-Box Searches with Partial STRIPS Models}

Our general hypothesis is that symbolic models can help to improve over black-box search methods that are guided with ML-based heuristics and/or policies. We assume a setting, where a practitioner starts with an already trained ML-based method and a black-box simulator, similar to the setting commonly used in Deep RL (DRL)~\cite{arulkumaran2017deep}. Our hypothesis is that, when ML-based black-box does not scale well, the practitioner can specify a partial STRIPS model and select a suitable search algorithm to lead to further scalability.  This hypothesis does not imply that, for a given ML-based method and partial STRIPS model, such an improvement will always manifest.  The empirical results in our evaluation show that such improvements are possible in black-box planning. In the rest of the section we answer questions crucial for proving our hypothesis: what ML-based methods we consider, how can partial STRIPS models be defined, and what search algorithms can be be used for combining the strengths of the ML-based methods and the heuristics obtained from the partial STRIPS models.

\subsection{ML-based methods}

RL methods focus on estimating one of two quantities.
One is \textit{value estimation}, that is the expected cost/reward of acting from a state.
The second one is \textit{action selection} where an action is chosen for a given state.
They are related, respectively, to the standard RL algorithms value-iteration and policy-iteration \cite{sutton-barto-2018}.
For instance, DRL algorithms for learning a policy might use ML for classifying which action should be applied in each state.
In such cases, the ML model returns an estimate of the likelihood of the actions.
If the estimation were perfect, maximizing the likelihood corresponds to an optimal policy for the problem.
When we are using a policy, we will assume that we can rank actions according to its likelihood.

We focus on black-box planning, where these are used to guide a best-first search, either selecting the state with best value or the node with highest likelihood. Therefore, in this work we refer to value estimation as \textit{heuristics}, and action selection as \textit{policies}.


\subsection{Partial STRIPS Models}

We assume that we are provided a black-box task, \bbtask, for which we have limited ML-based search
guidance. Our motivation is that practitioners would write down or extract a symbolic
model that represents part of the dynamic of the environment. Such a model would be
intentionally added, so practitioners have incentives to capture the weakness of their
ML-based method. To do so, they will provide a STRIPS task \ptask as well as a way of
mapping states from \bbtask to \ptask.

\begin{definition}[Partial Model]
Let $\bbtask = \langle S, A, \Sinit, \mathit{succ}, \mathit{goal} \rangle$ be a black-box planning task. A partial
model of \bbtask is a tuple $\langle \ptask, \sigma\rangle$ where $\ptask
= \langle \Facts, \Ops, \sigma(\Sinit), \Goal \rangle$ is a STRIPS task, and $\sigma$ is a
mapping $S \mapsto 2^F$ from states in \bbtask to states in \ptask.
\end{definition}

Partial models allow using any heuristic function that has been defined for STRIPS tasks
on searches for \bbtask. In particular, let $h_D$ be any heuristic for \ptask, we define a
heuristic for \bbtask as $h_{B} (s) = h_{D}(\sigma(s))$.

In principle, there are no requirements on the relation between \bbtask and \ptask. For example, the set of actions in \ptask can be completely different as that of \bbtask. If the preconditions and effects of actions in \bbtask are hard to formalize in STRIPS, one can model alternative actions encoding each original action as a sequence of STRIPS actions. But of course, the properties of $h_{B}$ greatly depend on the relation between \bbtask and \ptask. An interesting case is whenever \ptask models only part of the state, and the actions in \ptask expressing the same transition function as those in \bbtask. In that case, \ptask is a projection of \bbtask and the heuristic $h_{B}$ preserves the properties safeness, admissibility and consistency of the heuristic $h_D$ over \ptask~\cite{culberson-schaeffer-compint1998,edelkamp-ecp2001}.




\subsection{Algorithms}
\label{sec:algorithms}
We consider two search algorithms that can use both methods we have for guiding the search, the ML-based method and the heuristics from the partial STRIPS models:
\begin{enumerate}
\item \emph{Multi-heuristic best-first search}~\cite{helmert-jair2006} is a variation of \emph{Greedy Best-First Search (GBFS)} with two queues, which interleaves the corresponding queue to select for expansion its best node. We call this algorithm \textit{double-queue}.
\item GBFS with tie-breaking, which selects for expansion the best node according to the main heuristic. If more than one node has the best heuristic value, it uses the secondary heuristic for tie-breaking. 
\end{enumerate}
For dealing with policies, we use the concept of \emph{discrepancy}~\cite{harvey-ginsberg-ijcai1995}, as was defined by \citeauthor{karoui2007yields} (\citeyear{karoui2007yields}).
The method ranks the successors of a node  (starting from 0) in ascending order according to their heuristic value and/or the preferences of the policy. Then, the heuristic value of a node is the sum of the value of its parent plus its rank. Therefore, the value of each node is the sum of all ranks from the initial state to the node resulting on a path-dependent heuristic. This method was previously used in the context of bounded suboptimal search with good results~\cite{araneda-et-al-2021,greco-et-al-2022}.



\section{Methodology}

Even though our hypothesis is that symbolic models can help to improve over ML-based
methods in the general case where a complete declarative model is unknown, we evaluate it
on a controlled environment using multiple classical planning domains (see
Section~\ref{sec:domains}) for which a complete STRIPS model is already available. This
allows us to analyze the benefits from using partial models in a systematic way, using
several models per domain that include the ``original'' model that perfectly describes the
dynamics of the environment.

\subsection{Implementation}

All algorithms were implemented on the Fast Downward (FD) Planning System~\cite{helmert-jair2006}. To compute heuristics on partial STRIPS models, we implemented them as domain-dependent task transformations within the planner, allowing the computation of arbitrary heuristic functions on the partial model.

In our experiments, we use the FF heuristic~\cite{hoffmann-nebel-jair2001}, one of the most popular heuristics for satisficing planning due to its ability for guiding the search towards the goal in a broad class of domains~\cite{hoffmann-jair2005}. Note that our hypothesis is that ML-based methods can be improved by computing \emph{some} relaxation-based heuristic over a partial model. Therefore, our conclusions are not specifically tied to this heuristic and other planning heuristics might be more convenient for other domains.

\subsection{ML-based baseline}

We found it convenient to use HGN~\cite{shen-et-al-icaps2020}, a method that can produce a high-quality heuristic for specific domains and generalizes well to instances of different size. HGN
provides an estimate of the distance to the goal as a real, floating-point number. As
Fast Downward uses only integers, we take only $3$ decimals of precision, by multiplying
the result by $10^3$ and rounding to the nearest integer. We trained the heuristic using a
5-fold cross-validation with their default hyperparameters, i.e. hidden size = 32, bins=4
and learning rate = 0.001. We denote this heuristic as \hgn.

We also consider using HGN as a ``policy'', by considering that the policy recommends to take successors with lowest heuristic value. As mentioned above, we can handle this by using discrepancy, so that the evaluation of a node depends on the difference in heuristic value with other siblings. We denote this heuristic as \hgndisc.

In order to analyze how partial declarative models can be useful on scenarios where the
NN-based heuristics and/or policies are used in an out-of-distribution scenario, we train
HGN on different datasets. In all cases, our datasets consist of $100$ instances, which
are different from the instances used in our evaluation. The datasets do differ on the way
instances are generated. The default variant of our heuristics, \hgn and \hgndisc, are
trained on instances generated by the same generator used for obtaining the test set. The
training instances are typically smaller than the test instances, as is common practice in
learning approaches for planning. Other heuristics, which we will denote,
$\hgnd{{dataset}}$ or $\hgndiscd{{dataset}}$, use a ``biased'' training set,
where we vary the distribution of initial states and/or goals. In
Section~\ref{sec:domains}, we describe the different variants chosen for each domain.



\subsection{Metrics}

In our analysis, we focus on comparing the performance of all algorithms in terms of
number of expansions and do not focus on running time. This is desirable for several
reasons. First of all, in our experiments the time spent on the successor generation
and/or on computing the model-based heuristics such as FF is negligible to that of
evaluating NN-based heuristics such as HGN.  Therefore, in terms of runtime it is
sometimes desirable not to use the NN-based heuristics. However, we are interested in a
setting where using a ML-based heuristic or policy is desirable, e.g., because it is able
to capture large parts of the model that are subsymbolic and cannot be considered.
Furthermore, runtime measures could be influenced by a number of factors.  In practice,
different NN-based heuristics might have different computational cost, or specific
hardware can accelerate their computation.  Besides, whenever the black-box simulator is
based on NN predictions (e.g. as in model-based reinforcement learning), the overhead of
the successor generation will increase significantly, heavily limiting the amount of
nodes that can be expanded within a given time limit.

Therefore, we analyze how both sources of information, \hgn and \hff, can be combined in
order to reduce the search effort in terms of node expansions. Focusing on the number of
expansions enables a meaningful comparison when we evaluate not using the NN-based
heuristics, so we can assess the power of the partial STRIPS model isolated. And all the
conclusions obtained regarding the cases where using partial models can improve the
performance in terms of node expansions can be easily extrapolated to runtime, as the
computational effort of these heuristics is far from being a bottleneck.

\section{Benchmark Domains}
\label{sec:domains}

We consider three domains from the International Planning
Competition~\cite{mcdermott-aimag2000}. We selected domains that are non-homogeneous,
meaning that they have objects of different types so that it is natural to model only part
of the domain in a declarative fashion by specifying only a sub-set of the actions,
predicates and/or object types in PDDL.

\subsection{Logistics}

Our first domain is the IPC-00 version of Logistics. The task consists of delivering
packages from their initial location to their destination on a map that consists of
locations grouped in different cities. Across cities, packages must be transported by
airplanes, which are only allowed to move in between special locations where there is an
airport. Within each city, packages are transported via trucks.

Our partial model, called \textit{\logisticsone}, only considers transportation by airplane. In \logisticsone,
objects represent packages, cities, and airplanes, but there are no trucks and/or
locations within the cities. The goal is to have each package in the correct city. Mapping
states from the original task to the partial model is straightforward, by mapping the
current location of each package and/or airplane into the corresponding city. This models
a scenario where the exact location of each object is hard to determine in a symbolic
manner, but it is possible to determine a broad area where the object is located.

For training the HGN heuristic, the default dataset (\emph{hgn}) contains 100 instances with 2 to 4 packages, 2 to 5 cities, 2 to 4 airplanes and 2 to 5 locations in each city.
We also generate a biased dataset, called \textit{\logisticsdumbone}, where instances have the same size but all packages are placed in the same city in the initial state and goal, so it is never necessary to transport them by airplane. The resulting heuristic is expected to approximate the cost of moving trucks, but ignore all actions that involve the airplanes.

The test set contains 50 instances with 3 to 4 packages, 5 to 7 cities, 4 to 5 airplanes and 4 to 6 locations in each city.

\subsection{Grid}

In the IPC-00 grid domain, a robot can move along a grid transporting keys. The robot can
hold one key at a time. If a tile is locked, the robot has to carry a key of the
corresponding shape to unlock it. There are actions that move the robot, pick up or leave
keys at a given location, and unlock cells. The goal is to place certain keys at specific
locations.
For this domain, we consider these two partial domains:

\begin{enumerate}

\item \textit{\gridtwo}: We model only the position and movement of the robot, but no information
  regarding the keys is modelled, except which keys have already been dropped at their
  target. In this case, the planning heuristic estimates the distance the robot should traverse to
  visit all target locations where keys should be dropped, without having any information
  regarding the position of the keys. This could be the case in scenarios where a model of
  the environment of the robot is available but the position of the keys is not available
  in the symbolic model.

\item \textit{\gridthree}: We only model the position of the keys, but we do not model the position
  and/or movement of the robot. Therefore, each key can be grabbed and/or left at any
  position at any time. The heuristic guides the search towards states where more keys
  have been dropped at their destination. This kind of model could be useful in settings
  where the movement of the robot is complex to describe in a symbolic manner accurately.
\end{enumerate}

For training, we generate 100 instances of the following datasets, included two biased ones:
\begin{enumerate}
\item \textit{default (hgn)}: a grid size between 4x4 to 6x6, 1 to 2 locks and 1 to 3 keys.
\item \textit{\griddumbone}: similar to \textit{default}, but all instances have only two positions (grid size of 2x1 tiles) and one is locked. In the instances exists different keys. The robot has to move to the other location trying the correct key. In this case, the trained heuristic can learn that locked positions require to be unlocked by using keys.
\item \textit{\griddumbtwo}: similar to \textit{default}, but all instances have a single locked position, even though they may exist multiple keys. Trained heuristics with this data might be sensitive respect to the kind of grid where the robot moves.
\end{enumerate}
Finally, the test set contains 50 instances with grid size between 7x6 to 8x6, 2 to 3 locks and 2 to 4 keys.

\begin{table*}[t]
\centering
\begin{subtable}[h]{0.27\textwidth}
\begin{tabular}{l|rrr} \toprule
& none & $\hff_{\logisticsoneshort}$ & \hff \\ \midrule
none & - & 0 & \graytext{50} \\ \midrule
\hgnlogdumbone & 5 & \textbf{17} & \graytext{50} \\
\hgn & 44 & \textbf{50} & \graytext{50} \\
\midrule
\hgndisclogdumbone & 3 & 3 & \graytext{50} \\
\hgndisc & 50 & 50 & \graytext{50} \\
\end{tabular}
\caption{Logistics}
\label{table:eagergreedy-logistics}
\end{subtable}
\hfill
\begin{subtable}[h]{0.32\textwidth}
\begin{tabular}{l|rrrr} \toprule
 & none & $\hff_{\gridtwoshort}$ & $\hff_{\gridthreeshort}$ & \hff \\ \midrule
none & - & 34 & 50 & \graytext{50} \\ \midrule
\hgngriddumbone & 17 & 34 & 50 & \graytext{50} \\
\hgngriddumbtwo & 49 & 49 & 50 & \graytext{50} \\
\hgn & 50 & 50 & 50 & \graytext{50} \\
\midrule
\hgndiscgriddumbone & 33 & \textbf{49} & 50 & \graytext{50} \\
\hgndiscgriddumbtwo & 50 & 50 & 50 & \graytext{50} \\
\hgndisc & 50 & 50 & 50 & \graytext{50} \\
\end{tabular}
\caption{Grid}
\label{table:eagergreedy-grid}
\end{subtable}
\hfill
\begin{subtable}[h]{0.32\textwidth}
\begin{tabular}{l|rrrr} \toprule
& none & $\hff_{\woodtwoshort}$ & $\hff_{\woodoneshort}$ & \hff \\ \midrule
none & - & 2 & 19 & \graytext{50} \\ \midrule
\hgnwooddumbtwo & 14 & \textbf{24} & \textbf{24} & \graytext{50} \\
\hgn & 16 & \boldred{11} & \textbf{36} & \graytext{50} \\
\hgnwooddumbone & 26 & \textbf{33} & \textbf{40} & \graytext{50} \\
\midrule
\hgndisc & 10 & \textbf{11} & {14} & \graytext{49} \\
\hgndiscwooddumbone & 12 & \textbf{24} & \boldred{10} & \graytext{50} \\
\hgndiscwooddumbtwo & 17 & \textbf{23} & \textbf{18} & \graytext{49} \\
\end{tabular}
\caption{Woodworking-PD}
\label{table:eagergreedy-wood}
\end{subtable}
\caption{Coverage of GBFS combining variations of \hgn and \hff with the double-queue algorithm.  The
  top segment is coverage using only STRIPS models.  
  The left segment is coverage using only HGN.
  For comparison, the right segment is coverage using a complete STRIPS model.
  The middle segments are using HGN as
  a heuristic. The bottom ones are coverage using HGN as a policy. Variations of \hgn and
  \hgndisc indicate the data used for training. Variations of \hff indicate the partial
  STRIPS model. Cells in bold indicate those cases where the combination outperforms both FF and HGN in isolation. We highlight in red those cases where using a partial STRIPS model is detrimental. }
\label{table:eagergreedy-all}
\end{table*}

\subsection{Woodworking with Pickup and Delivery}

This is a variation of the IPC-11 domain Woodworking. The domain simulates the work in a
woodworking workshop where some wooden pieces are prepared by cutting boards and
varnishing, polishing or coloring the resulting pieces, using different tools, such as a
grinder, saw, glazer or spray. In our variant of this domain (henceforth Woodworking-PD),
there is additionally a road map. All tools are in a location (representing the workshop),
but the pieces of wood are scattered across multiple locations and they should be picked
up, brought to the workshop, and the prepared pieces should be delivered afterwards.  Thus, the domain
includes trucks to move the pieces of board. The original version of this domain has
action costs, but we consider a unit-cost version, as action costs are not supported by
the HGN heuristic. In our formulation, we only include one truck.
For this domain, we consider multiple partial models:
\begin{enumerate}
\item \textit{\woodone}: This is the original woodworking model, which ignores the pickup and
  delivery part.
\item \textit{\woodtwo}: This partial model ignores the woodworking part and considers that all
  objects are finished but in different places, locating each unfinished object at the workshop.
The tasks consists only of moving objects from one location to another.
\end{enumerate}

For training, we generate 100 instances of the following datasets, which included two biased ones that consider just the woodworking part or the pickup and delivery part.

\begin{enumerate}
\item \textit{default (hgn)}: the instances have 1 to 2 pieces, 2 to 6 locations, 3 to 5 machines and a wood factor (amount of wood with respect to the minimum needed) of 1.4 to 1.0.
\item \textit{\woodworkingdumbone}: similar to \textit{default}, but with only one location, and all pieces and machines are in there. For that reason, the goal can be achieved without moving the truck, similar to the original woodworking domain.

\item \textit{\woodworkingdumbtwo}: similar to \textit{default}, but for achieving the goal, it is necessary to move the corresponding pieces to a location and not consider processing the pieces (such as paint, varnish, polish, etc). Thus, the goal can be achieved by just moving the truck.

\end{enumerate}

Finally, for testing, we generated a dataset with 50 instances with 2 to 3 parts, 3 to 8 locations, and a wood factor of 1.4 to 1.0.

\begin{figure*}[ht]
    \centering
    \begin{subfigure}{0.3\textwidth}
    \includegraphics[width=\textwidth]{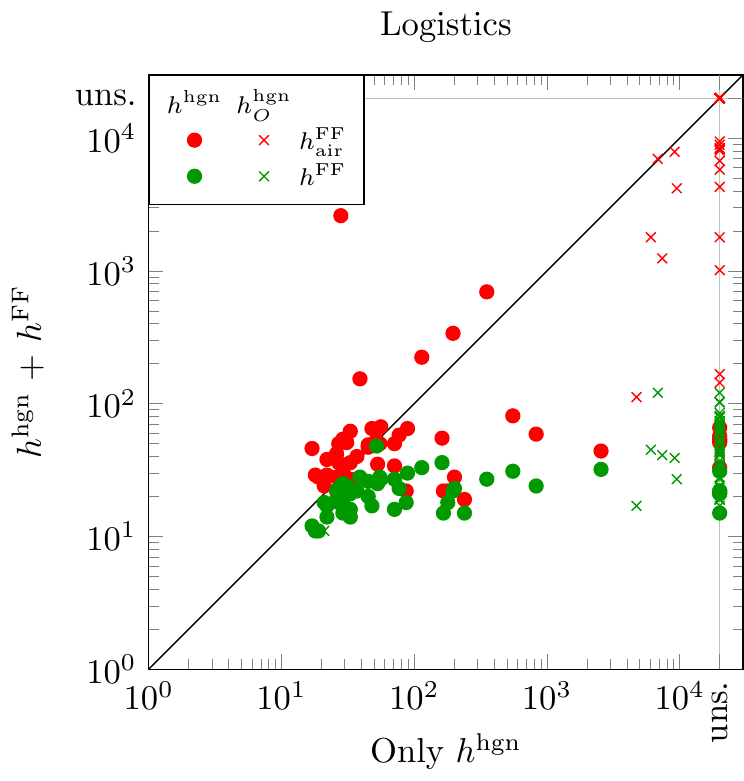}
    \end{subfigure}
    \qquad
        \begin{subfigure}{0.3\textwidth}
    \includegraphics[width=\textwidth]{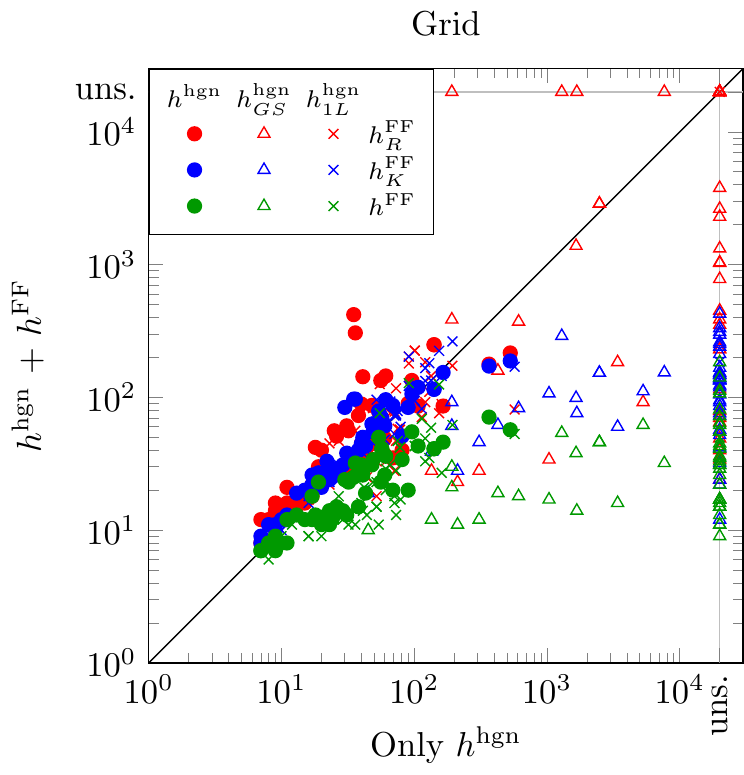}
    \end{subfigure}
    \qquad
    \begin{subfigure}{0.3\textwidth}
    \includegraphics[width=\textwidth]{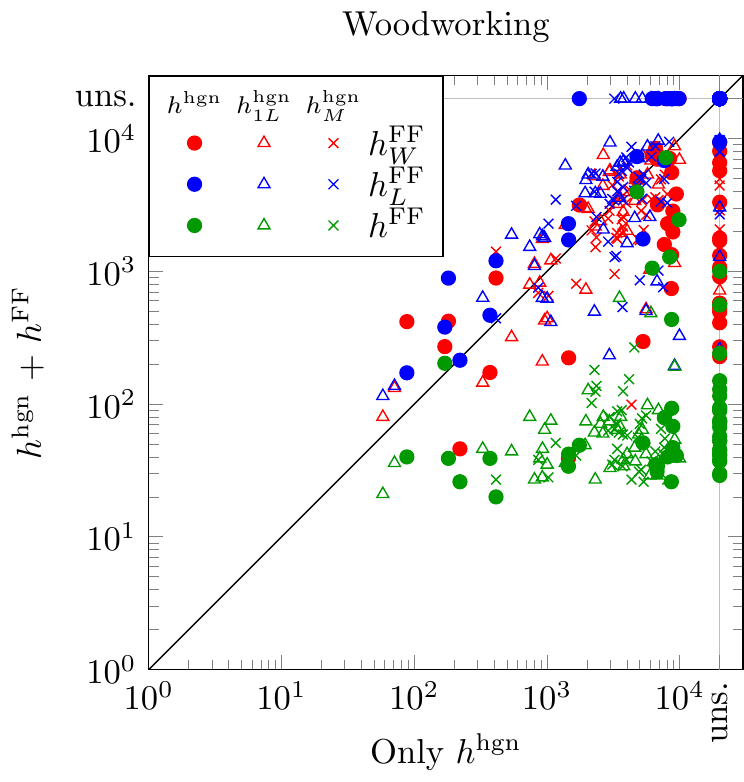}
    \end{subfigure}

    \bigskip

    \begin{subfigure}{0.3\textwidth}
    \includegraphics[width=\textwidth]{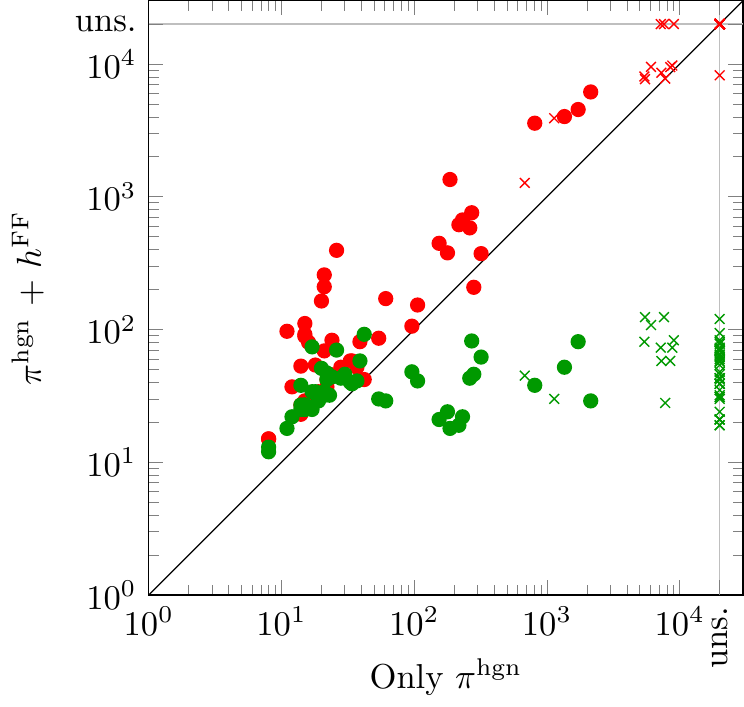}
    \end{subfigure}
    \qquad
        \begin{subfigure}{0.3\textwidth}
    \includegraphics[width=\textwidth]{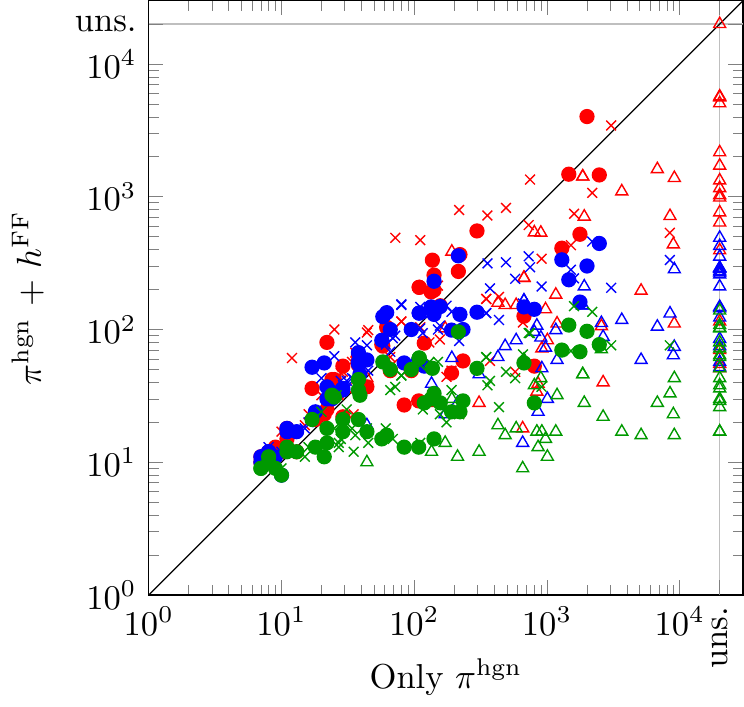}
    \end{subfigure}
    \qquad
    \begin{subfigure}{0.3\textwidth}
    \includegraphics[width=\textwidth]{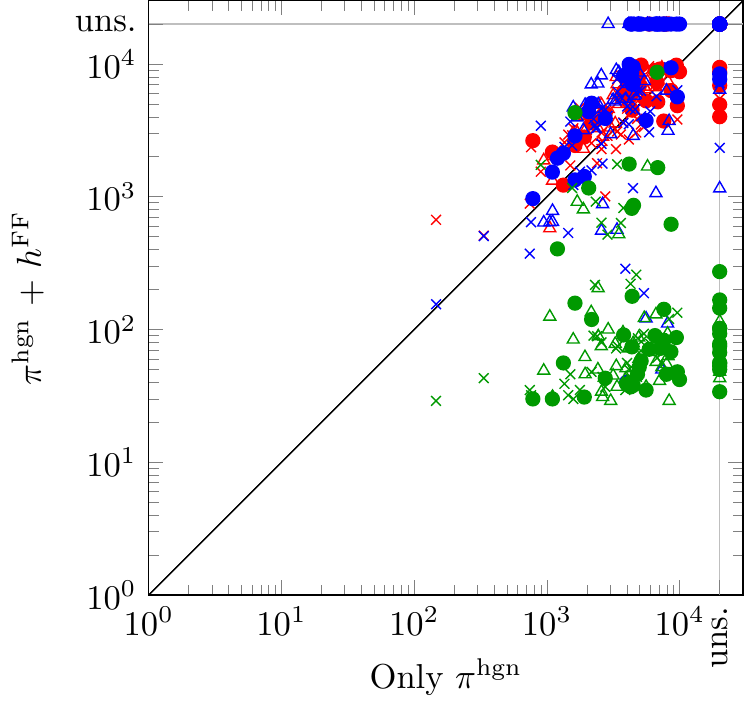}
    \end{subfigure}
    \caption{Expansions of only HGN against using both HGN and \hff with different partial
      models. In the top row, HGN is used as heuristic (e.g. \hgn) and in the bottom row, it is used as a policy (e.g. \hgndisc). We distinguish different variants of HGN trained with different
      datasets with shapes, and different partial STRIPS models with colors. }
    \label{fig:scatter}
\end{figure*}

\begin{figure}[t]
\centering
\includegraphics[width=0.67\columnwidth]{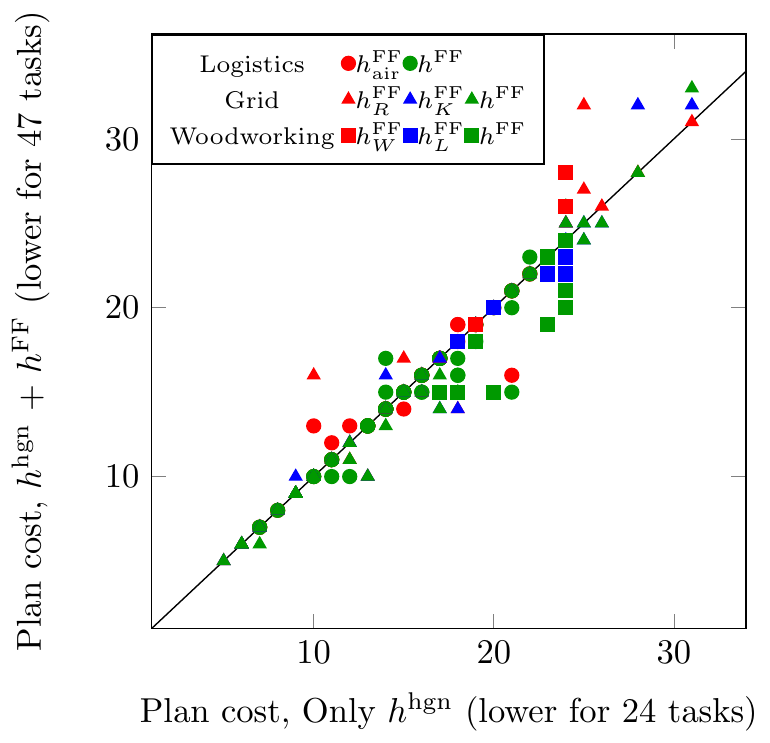}
\caption{Plan cost when using only \hgn against using the double queue method with \hgn and partial-model heuristics, including the \hff.}
\label{fig:costs}
\end{figure}

\section{Evaluation}

We report experiments on how partial STRIPS models improve over ML baselines. All
experiments were run with a time limit of 120 minutes, a memory limit of 9GB, and were
executed on a cluster with Intel Xeon Gold 6126 CPU with hyper-threading enabled and
around 30 tasks in parallel using Downward lab \cite{seipp-et-al-zenodo2017}.
All our code, benchmarks and data are publicly available \cite{codeofthispaper}.

For the evaluation, we use $50$ instances in each domain, all different from those used for
training the heuristics. In all our experiments, we limit the number of expansions by all
configurations to \numprint{10000}, the number of expansions that all configurations could
complete under our time limit.

\subsection{Guiding Search with Partial STRIPS Models}

Table~\ref{table:eagergreedy-all} reports the coverage when using different kinds of HGN
heuristics (trained with the original dataset or with biased ones), and FF heuristics
(using either the full or partial STRIPS models).

The top segment of each table is the coverage when using only the planning heuristic over
partial STRIPS models. As a reference for comparison,
we include full STRIPS model combined with the FF heuristics (\hff).
Without having access to any learned heuristic, it solves all instances under \numprint{10000} expansions.
That offers an upper bound on how partial STRIPS model can help on domains that are not
modeled completely.


The results support our hypothesis that partial STRIPS model can be useful to enhance the
guidance of weak ML-based heuristics and policies. In particular, coverage often increases
compared to the variant that uses only the HGN heuristics. There are only two cases,
marked in {red}, where there is a slight decrease in coverage.

Nevertheless, as expected, the quality of the guidance of a partial model heavily depends
on the part of the problem that has been modelled.  In some cases guidance is as good as
with the perfect model. For example, in Grid-\gridthree, by modelling only the position of
the keys and ignoring the robots, the search is guided towards relevant sub-goals
(picking-up necessary keys and/or dropping them down at their locations).
In other cases, however, guidance is more limited. For example, the partial model
Logistics-\logisticsone, ignores completely all locations within a city so it assigns a
heuristic value of 0 to all states where the packages are in the right city, regardless of
whether the goal has been fulfilled. This causes huge heuristic plateaus and hence the
coverage is 0 when used alone. The great news is that, even in such cases, the heuristic
can be a good complement for other ML-based heuristics. As shown in the results from
Table~\ref{table:eagergreedy-all}, coverage increases with respect to using only \hgn. A likely
reason is that $\hff_{\logisticsoneshort}$ helps to guide the search until having all
packages in the right city, and then \hgndisc/\hgndisclogdumbone can help to achieve the
goal from there.

Figure~\ref{fig:scatter} shows a more detailed view on the comparison on search effort in
terms of expansions of only using HGN with respect to combining HGN and FF under different
partial models. In general, combining HGN and partial STRIPS leads to more points under
the diagonal, validating our hypothesis. This is partially due to the way heuristics are
combined, by having two separate open lists one for each heuristic, which keeps the
overhead small even when the heuristic provided by the partial model is not accurate
(e.g. in Grid when using the \gridtwo partial model, $\hff_{\gridtwoshort}$, and the HGN
trained on a non-biased dataset). On the other hand, partial models can reduce search
effort significantly by several orders of magnitude, whenever they model a part of the
domain that was not correctly captured by the HGN heuristic.  We remark again that, the
computation of the FF heuristic does not have a huge overhead with respect to HGN, and
hence the advantages in terms of expanded nodes also carry to search time.

Regarding the cost of the plans found by the different approaches, we observe that there
is not a very high variance. As shown in Figure~\ref{fig:costs}, even when using the best
HGN variant, plan costs are slightly better when using the partial-model heuristics
(finding better plans in 47 tasks compared to 24). When excluding the full model
configurations (\hff), this is tied (improving in 19 tasks compared to 18), and if the
biased HGN variants are considered the advantage is clearer (132 tasks compared to
55). Overall, results regarding plan cost seem to correlate with tasks solved, so that
more informative heuristics also find (slightly) better plans.






\subsection{Other ways of combining heuristics}
\begin{figure*}[t]
    \centering
    \begin{subfigure}{0.49\textwidth}
    \begin{tikzpicture}
  \begin{axis}[
      xlabel={Expansions},
      ylabel={Coverage},
      scale only axis,
      xmode=log,
      xmax = 10000,
      unbounded coords=jump,
      legend pos= north west,
      legend columns = 1,
      legend style={fill=white, fill opacity=0.4, draw opacity=1,text opacity=1},
      width=0.9\linewidth,
      height=0.6\linewidth,
      cycle list ={
            thick    \\
            thick, blue, dotted\\%
            thick, red, loosely dashed\\%
            thick, green, dashdotted \\%
            thick, orange, dashdotdotted\\%
            thick, yellow, densely dotted\\%
            thick, blue, densely dotted\\%
            thick, green, densely dotted\\%
            thick, orange, densely dotted\\%
            thick, black, densely dotted\\%
            thick, red, densely dotted\\%
            thick, densely dashed\\%
            thick, blue, densely dotted\\%
      }
      ]

\addplot table [x=expansions, y=none@none@fft2@eagergreedy] {cumulative_plots/woodworking10.dat};
\addplot table [x=expansions, y=normal@hgndumb2@none@eagergreedy] {cumulative_plots/woodworking10.dat};
\addplot table [x=expansions, y=normal@hgndumb2@fft2@eagergreedy] {cumulative_plots/woodworking10.dat};
\addplot table [x=expansions, y=normal@hgndumb2@fft2@tiebreaking] {cumulative_plots/woodworking10.dat};
\addplot table [x=expansions, y=normal@hgndumb2@fft2@tiebreakingreverse] {cumulative_plots/woodworking10.dat};

  \end{axis}
\end{tikzpicture}
    \caption{Woodworking-PD $\hff_{\woodtwoshort}$ and \hgnwooddumbtwo}
    \label{fig:plot-woodworking10-normal-hgn2-fft2}
    \end{subfigure}
\hfill
    \begin{subfigure}{0.49\textwidth}
    \begin{tikzpicture}
  \begin{axis}[
      xlabel={Expansions},
      ylabel={Coverage},
      scale only axis,
      xmode=log,
      xmax = 10000,
      unbounded coords=jump,
      legend pos= north west,
      legend columns = 1,
      legend style={fill=white, fill opacity=0.4, draw opacity=1,text opacity=1},
      width=0.9\linewidth,
      height=0.6\linewidth,
      cycle list ={
            thick    \\
            thick, blue, dotted\\%
            thick, red, loosely dashed\\%
            thick, green, dashdotted \\%
            thick, orange, dashdotdotted\\%
            thick, yellow, densely dotted\\%
            thick, blue, densely dotted\\%
            thick, green, densely dotted\\%
            thick, orange, densely dotted\\%
            thick, black, densely dotted\\%
            thick, red, densely dotted\\%
            thick, densely dashed\\%
            thick, blue, densely dotted\\%
      }
      ]

\addplot table [x=expansions, y=none@none@fft1@eagergreedy] {cumulative_plots/woodworking25.dat};
\addplot table [x=expansions, y=discrank@hgncomplete@none@eagergreedy] {cumulative_plots/woodworking25.dat};
\addplot table [x=expansions, y=discrank@hgncomplete@fft1@eagergreedy] {cumulative_plots/woodworking25.dat};
\addplot table [x=expansions, y=discrank@hgncomplete@fft1@tiebreaking] {cumulative_plots/woodworking25.dat};
\addplot table [x=expansions, y=discrank@hgncomplete@fft1@tiebreakingreverse] {cumulative_plots/woodworking25.dat};

  \end{axis}
\end{tikzpicture}
    \caption{Woodworking-PD $\hff_{\woodoneshort}$ and \hgndisc}
    \end{subfigure}

    \medskip

    \begin{subfigure}{0.49\textwidth}
    \begin{tikzpicture}
  \begin{axis}[
      xlabel={Expansions},
      ylabel={Coverage},
      scale only axis,
      xmode=log,
      xmax = 10000,
      unbounded coords=jump,
      legend pos= south east,
      legend columns = 1,
      legend style={fill=white, fill opacity=0.4, draw opacity=1,text opacity=1},
      width=0.9\linewidth,
      height=0.6\linewidth,
      cycle list ={
            thick    \\
            thick, blue, dotted\\%
            thick, red, loosely dashed\\%
            thick, green, dashdotted \\%
            thick, orange, dashdotdotted\\%
            thick, yellow, densely dotted\\%
            thick, blue, densely dotted\\%
            thick, green, densely dotted\\%
            thick, orange, densely dotted\\%
            thick, black, densely dotted\\%
            thick, red, densely dotted\\%
            thick, densely dashed\\%
            thick, blue, densely dotted\\%
      }
      ]

\addplot table [x=expansions, y=none@none@fft2@eagergreedy] {cumulative_plots/grid2.dat};
\addplot table [x=expansions, y=normal@hgncomplete@none@eagergreedy] {cumulative_plots/grid2.dat};
\addplot table [x=expansions, y=normal@hgncomplete@fft2@eagergreedy] {cumulative_plots/grid2.dat};
\addplot table [x=expansions, y=normal@hgncomplete@fft2@tiebreaking] {cumulative_plots/grid2.dat};
\addplot table [x=expansions, y=normal@hgncomplete@fft2@tiebreakingreverse] {cumulative_plots/grid2.dat};

  \end{axis}
\end{tikzpicture}
    \caption{Grid $\hff_{\gridtwoshort}$ and \hgn}
    \label{fig:plot-grid2-bestrank-hgn1-fft2}
    \end{subfigure}
\hfill
    \begin{subfigure}{0.49\textwidth}
    \begin{tikzpicture}
  \begin{axis}[
      xlabel={Expansions},
      ylabel={Coverage},
      scale only axis,
      xmode=log,
      xmax = 10000,
      unbounded coords=jump,
      legend pos= south east,
      legend columns = 1,
      legend style={fill=white, fill opacity=0.4, draw opacity=1,text opacity=1},
      width=0.9\linewidth,
      height=0.6\linewidth,
      cycle list ={
            thick    \\
            thick, blue, dotted\\%
            thick, red, loosely dashed\\%
            thick, green, dashdotted \\%
            thick, orange, dashdotdotted\\%
            thick, yellow, densely dotted\\%
            thick, blue, densely dotted\\%
            thick, green, densely dotted\\%
            thick, orange, densely dotted\\%
            thick, black, densely dotted\\%
            thick, red, densely dotted\\%
            thick, densely dashed\\%
            thick, blue, densely dotted\\%
      }
      ]

\addplot table [x=expansions, y=none@none@fft2@eagergreedy] {cumulative_plots/grid26.dat};
\addlegendentry{\scriptsize Only FF}
\addplot table [x=expansions, y=discrank@hgncomplete@none@eagergreedy] {cumulative_plots/grid26.dat};
\addlegendentry{\scriptsize Only HGN}
\addplot table [x=expansions, y=discrank@hgncomplete@fft2@eagergreedy] {cumulative_plots/grid26.dat};
\addlegendentry{\scriptsize Double queue}
\addplot table [x=expansions, y=discrank@hgncomplete@fft2@tiebreaking] {cumulative_plots/grid26.dat};
\addlegendentry{\scriptsize FF, tiebreaking HGN}
\addplot table [x=expansions, y=discrank@hgncomplete@fft2@tiebreakingreverse] {cumulative_plots/grid26.dat};
\addlegendentry{\scriptsize HGN, tiebreaking FF}

  \end{axis}
\end{tikzpicture}
    \caption{Grid $\hff_{\gridtwoshort}$ and \hgndisc}
    \label{fig:plot-grid26-bestrank-hgn-fft3}
    \end{subfigure}
    \caption{Cumulative coverage by the number of expanded nodes under different variants of \hff, \hgn and \hgndisc.}
    \label{fig:cumulativeplots}
\end{figure*}
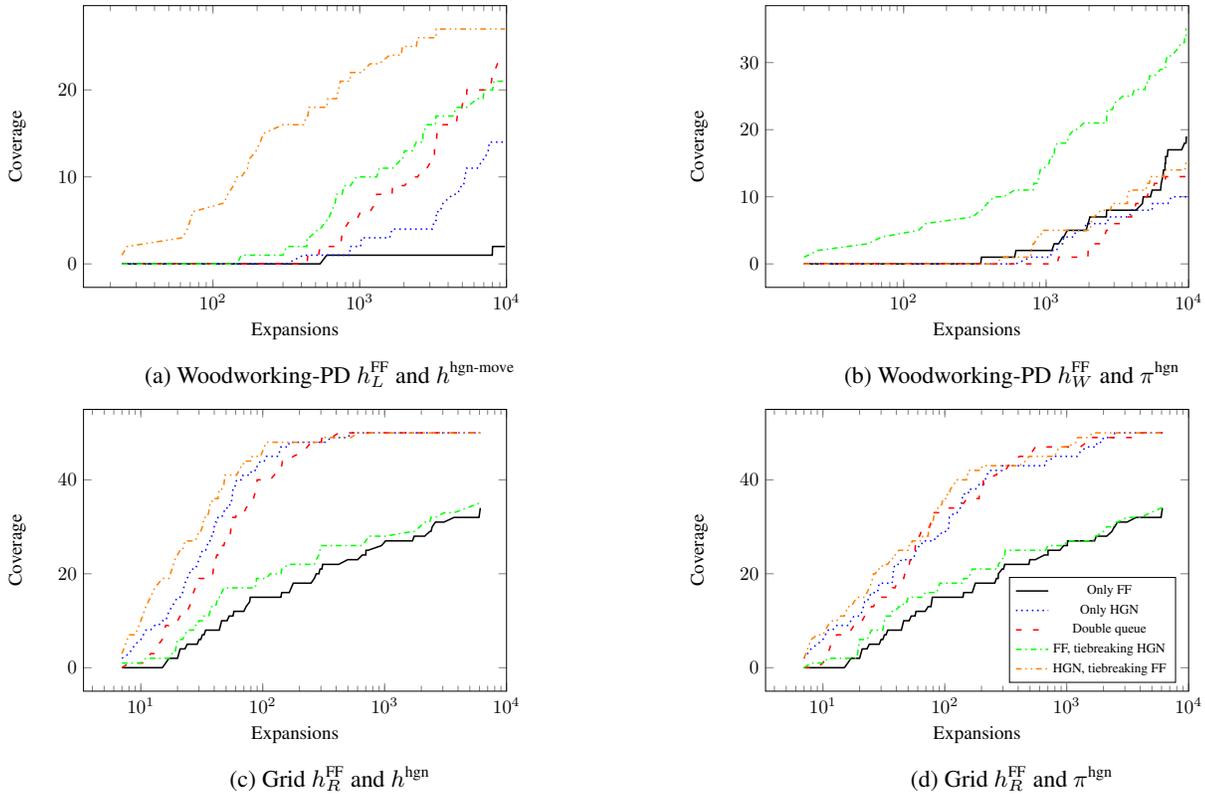

Table~\ref{table:eagergreedy-all} focuses exclusively on the performance using the double-queue algorithm. In general, this was the most suitable way for combining both heuristics. However, it is not always the best. Table~\ref{table:bestalgo-wood-rank} shows the cases where the tie-breaking combination mentioned in section~\ref{sec:algorithms} performed better in the Woodworking-PD domain. We observe that, especially with \hgndisc, the double-queue algorithm can be outperformed by tie-breaking. Typically, it is better to choose as primary heuristic the one that performs best in isolation, so in this case we used \hff as primary heuristic, breaking ties with \hgn.


Figure~\ref{fig:cumulativeplots} shows the cumulative coverage by expanded nodes of the
different algorithms. Each plot presents a specific configuration: (a) and (c) shows the
performance of \hgn; (b) and (d) of \hgndisc. Subfigure~\ref{fig:cumulativeplots}(a) shows
that, combining with the partial model, Woodworking-PD~\woodtwo outperforms all other
configurations and dramatically increases the number of solved instances. On the other
hand, (b) shows that FF tiebreaking with \hgndisc outperforms a double-queue algorithm. In
both, using some form of a combination of heuristics is better than using the only one. In
the Grid domain, we observe that using \hgn and \hgndisc, the three configurations that
combine the heuristics outperform the use of only one and work similarly.

However, as shown in table~\ref{table:eagergreedy-all} and Figure~\ref{fig:scatter}, there are situations where the algorithms we considered did not manage to improve over the HGN baseline.

\begin{table}
\centering
\begin{tabular}{l|rrrr} \toprule
& none & $\hff_{\woodtwoshort}$ & $\hff_{\woodoneshort}$ & \hff \\ \midrule
none &  & 2 & 19 & \graytext{50} \\ \midrule
\hgnwooddumbtwo & 14 & 24+3 & 24+5 & \graytext{50} \\
\hgn & 16 & 11+4 & 36+3 & \graytext{50} \\
\hgnwooddumbone & 26 & 33+1 & 40 & \graytext{50} \\
\midrule
\hgndisc & 10 & 11+1 & 14\textbf{+21} & \graytext{49} \\
\hgndiscwooddumbone & 12 & 24 & 10\textbf{+22} & \graytext{50} \\
\hgndiscwooddumbtwo & 17 & 23 & 18+5 & \graytext{49+1} \\
\end{tabular}
\caption{Woodworking-PD. Additional coverage by using an alternative algorithm vs table~\ref{table:eagergreedy-wood}. See legend in that table. Bold numbers show using using an alternative algorithm lead to a new case where the combination outperforms both HGN and STRIPS models in isolation.}
\label{table:bestalgo-wood-rank}
\end{table}

\section{Related work}

Planning as heuristic search approaches~\cite{bonet-geffner-aij2001} use the STRIPS model to relax or transform the state space and guide the search, either by estimating the distance from each state to the goal, or by computing preferences over which actions or state to explore first \cite{richter-helmert-icaps2009}. These last two ideas are related to the notions used in RL we mention in Section~\ref{sec:algorithms}: value-estimation and action-selection \cite{silver-et-al-nature2016}.

Beyond RL, ML sequence and structured prediction aims to produce sequences with high likelihood. They can be seen as policy estimations as the likelihoods are calculating using an internal state, the past decisions and classifying over the next possible tokens in the sequence or structure. For instance, \cite{scholak-et-al-2021} reports using a pre-trained language model to map natural language questions into syntactically correct SQL queries. Typically, beam search is used to find for a sequence with high aggregated likelihood. This case includes the widely used large language models \cite{vaswani2017attention}.

There has been work on comparing learned heuristics with classical planning heuristics~\cite{ferber-et-al-ecai2020}.
It is  known that declarative models enable efficient computation of heuristics that are informative and time-efficient~\cite{geffner-bonet-2013}.
As such, when a full model is available those heuristics can typically outperform NN heuristics even when training in a fixed task~\cite{ferber-et-al-ecai2020}. However, when a full model is not available, it is not clear if and how relaxation-based heuristics can be used.

Previous work has used symbolic information to improve the behaviour of RL algorithms.
For instance, \citet{lee-et-all-icaps2021wsprl} use planning actions to define RL options to be used at training, and \citet{illanes2020symbolic} use symbolic plans for the task to improve the learning stage.
Both \citet{giacomo-et-al-2019} and \citet{icarte-et-al-2022} use LTL for specifying RL rewards.
On the other hand, \citet{alshiekh-et-al-2018} enforce LTL formulas during training and inference so the actions taken are safe.
On the planning side, there has been work on obtaining or improving planning heuristics using ML \cite{yoon-et-al-icaps2006,virseda-et-al-icaps2013wspal,karia-srivastava-aaai2021,ferber-et-al-icaps2021wsprl,toyer-et-al-aaai2018}.
Some others have worked on using DRL for generalized planning for solving planning instances in a given domain \cite{rivlin-et-al-icaps2020wsprl}.
%
In contrast with these directions, our work emphasizes modifying the behaviour after the ML model has been trained, amortizing the training cost.
Moreover, our contributions suggest new ways for faster adaptation to new specific domains as far as part of the high level structure can be expressed as a (partial) STRIPS model.

Modelling planning problems is studied in knowledge engineering \cite{vaquero-et-al-2013}. Recent efforts have looked at obtaining planning models from source code using annotations \cite{katz-et-al-2018}.

Model-lite planning approaches~\cite{kambhampati-aaai2007,weber-bryce-2011,zhuo-kambhampati-ai2017} have considered incomplete planning models in the past, e.g., for the generation of robust plans. Instead, we explore how to use them to enhance existing ML-based systems.

\section{Discussion}



Pure RL methods can generalize well when the training data covers well-enough the distribution of trajectories~\cite{sutton-barto-2018}. On the other hand, it is common to find that RL models show weak generalization when they rely on passively collected data. In this case, the most common next steps are tuning the ML algorithm or collecting more data based on insights of error analysis. In this work we explored a complementary direction: improve the scalability by turning such insights into a symbolic model, exploiting classical planners which naturally deal well with generalization. We showed that, even when a full model is not available, classical
planning heuristics can complement well ML-based heuristics. Typically, more accurate models lead to better search guidance. But even models
that disregard large parts of the problem can be very beneficial.

In general, the actions returned by a classical planner can be executed safely and reach
the goal as far as the model correctly represents the real state space. However, classical
planners are also used in cases where this not the case. For instance, simple classes of
full-observable non-deterministic planning can be tackled by assuming a deterministic
problem and replanning when the execution fails~\cite{yoon-et-al-icaps2007}. While this
work is about black-box planning, it can also be seen as using planners while associating
planning states with external states. Complementing ML-based methods with model-based
heuristics could improve the impact of planning research, and lead to a deeper
understanding of the relationship between learning and reasoning.  For instance, one could
start with a trained policy where states are partially symbolic, and associate that part
of the state to an STRIPS model. In turn, this combination could offer an stronger
baseline for further research on DRL.

As future work, there are many directions worth pursuing.
We would like to test our methods using other ML models and in other domains.
For instance, our approach would be directly applicable on domains where the dynamics of the black-box simulator cannot be expressed in STRIPS, e.g., with non-symbolic states or continuous variables.

Another question is what are good partial STRIPS models. In
some cases, the ML models and the partial STRIPS model might be focusing on different
aspects of the problem, which could be in contradiction.
For instance, as training ML aims to minimize the expected cost, we might want to use a STRIPS model to avoid taking actions leading to undesirable states, leading the search towards more secure zones.
In principle, this would make the plans more reliable but we have not explored the overall behaviour of our algorithms in such cases.
One way to study this issue is to consider ML models trained
on data about less risky scenarios, while the partial STRIPS model focus on safety issues.

Finally, our tie-breaking configurations commit exclusively into the preferences of the
first heuristic.  Hence, it would interesting to explore other ways of balancing both heuristics, such as focal-search based algorithms~\cite{greco-baier-2021, greco-et-al-2022, KFSgreco-et-al-2022}.

\subsection*{Acknowledgements}
We would like to thank William Shen and Florian Gei$\ss$er for sharing the code of their implementation of the hypergraph networks (HGN) in Fast Downward. Matias Greco was supported by the National Agency for Research and Development (ANID) / Doctorado Nacional / 2019 - 21192036. Matias Greco and Jorge Baier thank to Centro Nacional de Inteligencia Artificial CENIA, FB210017, BASAL, ANID.

{
\bibliography{bib/abbrv,bib/literatur,bib/ref,bib/crossref}
}
\appendix







\end{document}